\pdfoutput=1
\documentclass[10pt,twocolumn,letterpaper]{article}

\usepackage{cvpr}
\usepackage{times}
\usepackage{epsfig}
\usepackage{graphicx}
\usepackage{amsmath}
\usepackage{amssymb}
\usepackage{color}
\usepackage[titletoc,toc,title]{appendix}
\usepackage{url}

\usepackage{breakurl}
\usepackage[final]{pdfpages}

\usepackage{color}
\usepackage[mathscr]{eucal}
\usepackage{amsbsy}
\usepackage{bm}
\usepackage{fixltx2e}
\MakeRobust{\overrightarrow}
\usepackage{booktabs}
\usepackage{mathtools}

\usepackage{amssymb}
\usepackage{amsmath,amsthm}
\usepackage{graphicx}
\usepackage{graphics}
\usepackage{color}
\usepackage{xspace}
\usepackage{bbm}
\usepackage{psfrag}
\usepackage{algorithmicx}
\usepackage{algorithm}
\usepackage{algpseudocode}
\usepackage{multirow}
\usepackage{array}
\usepackage{url}
\usepackage[normalem]{ulem}
\usepackage{floatflt,setspace}
\usepackage{algcompatible}

\newtheorem{theorem}{Theorem}
\usepackage{stackengine}

\algnewcommand\INPUT{\item[\textbf{Input:}]}%
\algnewcommand\OUTPUT{\item[\textbf{Output:}]}%

\usepackage{graphicx}
\usepackage{caption}
\usepackage{subcaption}
\usepackage{amsmath}
\usepackage{amssymb}
\usepackage{array}
\usepackage{booktabs}

\usepackage[pagebackref=true,breaklinks=true,letterpaper=true,colorlinks,bookmarks=false]{hyperref}

\newcommand{\red}[1]{{\color{red} #1}}

\newcommand{\construct}{\textbf{construct}}

\newcommand{\merge}{\textbf{merge}}
\newcommand{\U}{\mathscr{U}}
\newcommand{\V}{\mathscr{V}}
\newcommand{\decomp}{\textbf{decomp}}

\cvprfinalcopy 

\def\cvprPaperID{3989} 

\ifcvprfinal\fi
\begin{document}
\title{Wide Compression: Tensor Ring Nets}

\author{Wenqi Wang\\
Purdue University\\
{\tt\small wang2041@purdue.edu}
\and
Yifan Sun\\
Technicolor Research\\
{\tt\small ysun13@cs.ubc.ca}
\and
Brian Eriksson\\
Adobe\\
{\tt\small eriksson@adobe.com}
\and
Wenlin Wang\\
Duke University\\
{\tt\small wenlin.wang@duke.edu}
\and
Vaneet Aggarwal\\
Purdue University\\
{\tt\small vaneet@purdue.edu}
}

\maketitle

\begin{abstract}
Deep neural networks have demonstrated state-of-the-art performance in a variety of real-world applications.  In order to obtain performance gains, these  networks have grown larger and deeper, containing millions or even billions of parameters and over a thousand layers.  The trade-off is that these large architectures require an enormous amount of memory, storage, and computation, thus limiting their usability.  
Inspired by the recent tensor ring factorization, we introduce  Tensor Ring Networks (TR-Nets), which  significantly compress both the fully connected layers and the convolutional layers of deep neural networks.   
Our  results show that our TR-Nets approach {is able to compress LeNet-5 by $11\times$ without losing accuracy}, and can compress  the state-of-the-art Wide ResNet
by  $243\times$ with only 2.3\% degradation in {Cifar10 image classification}.
Overall, this   compression scheme shows promise in scientific computing and deep learning, especially for emerging  resource-constrained devices such as  smartphones, wearables, and IoT devices.
\end{abstract}
\vspace{-.2in}
\section{Introduction}
\label{sec:intro}

Deep neural networks have made significant improvements in a variety of applications, including recommender systems~\cite{van2013deep,zhang2017deep}, time series classification~\cite{wang2016earliness}, nature language processing~\cite{collobert2008unified,graves2013speech,wang2017topic}, and image and video recognition~\cite{yang2017tensor}.  
These accuracy improvements require developing deeper and deeper networks, evolving
from AlexNet ~\cite{krizhevsky2012imagenet} (with $P = 61$ M parameters), VGG19 ~\cite{simonyan2014very} ($P = 114$ M), and GoogleNet ~($P = 11$ M)  \cite{szegedy2015going}, to 32-layer ResNet ($P = 0.46$ M) ~\cite{he2016deep,he2016identity},  28-layer WideResNet~\cite{zagoruyko2016wide} ($P = 36.5$ M), and DenseNets~\cite{huang2016densely}.  
Unfortunately, with each evolution in architecture comes a significant increase in the number of model parameters. 

On the other hand, many modern use cases of deep neural networks are for resource-constrained devices, such as mobile phones \cite{kim2015compression}, wearables and IoT devices \cite{lane2015early}, etc. In these applications,  storage, memory, and test runtime complexity are extremely limited in resources, and compression in these areas is thus essential.

After prior work~\cite{ba2014deep} observed redundancy in trained neural networks, a useful area of research has been compression of network layer parameters ({\em e.g.,}~\cite{chen2015compressing,han2015learning,han2015deep,denton2014exploiting}).  While a vast majority of this research has been focused on the compression of fully connected layer parameters, the latest deep learning architectures are almost entirely dominated by convolutional layers.  For example, while only 5\% of AlexNet parameters are from convolutional layers, over 99\% of Wide ResNet  parameters are from convolutional layers.  This necessitates new techniques that can factorize and compress the multi-dimensional tensor parameters of convolutional layers.

We propose compressing deep neural networks using
\emph{Tensor Ring (TR) factorizations}~\cite{zhao2016tensor},
which can be viewed as a generalization of a single Canonical Polyadic (CP) decomposition \cite{hitchcock1927expression,kolda2009tensor,ashraphijuo2017approximation}, with two extensions:
	\begin{enumerate}
		\vspace{-.06in}
		\item the outer vector products are generalized to matrix products, and 
		\vspace{-.1in}
		\item the first and last matrix are additionally multiplied along their outer edges, forming a ``ring" structure.
		\vspace{-.06in}
	\end{enumerate}
The exact formulation is described in more detail in Section \ref{sec:model}.
Note that this is also a generalization of the \emph{Tensor Train factorization}~\cite{oseledets2011tensor}, which only includes the first extension.
This is inspired by previous results in image processing \cite{wang2017efficient}, which demonstrate that this general factorization technique is extremely expressive, especially in preserving spatial features.

Specifically, we introduce Tensor Ring Nets (TRN), in which layers of a deep neural network  are compressed using tensor ring factorization. 
For fully connected layers, we compress the weight matrix, and investigate different merge/reshape orders to minimize real-time computation and memory needs. 
For convolutional layers, we carefully compress the filter weights such that we do not distort the spatial properties of the mask. 
Since the mask dimensions are usually very small ($5\times 5$, $3\times 3$ or even $1\times 1$) 
we do not compress along these dimensions at all, and instead compress along the input and output channel dimensions.  

To verify the expressive power of this formulation, we train several compressed networks.
First, we train LeNet-300-100 and LeNet-5 \cite{lecun1998gradient} on the MNIST dataset, {compressing LeNet-5 by $11\times$ without degradation and achiving $99.31\%$ accuracy, and compressing LeNet-300-100 by $13\times$ with a degrading of only $0.14\%$ (obtaining overall accuracy of $97.36\%$)}.  
Additionally, we examine the state-of-the-art 28-layer Wide-ResNet \cite{zagoruyko2016wide} on Cifar10, and find that TRN can be used to effectively compress the Wide-ResNet by $243 \times$ with only $2.3$\% decay in performance, obtaining $92.7\%$ accuracy.
The compression results demonstrates the capability of TRN to compress state-of-the-art deep learning models for new resources constrained applications. 

Section~\ref{sec:relWork} discusses related work in neural network compression.
The compression model is introduced in  Section~\ref{sec:model}, which discusses general tensor ring factorizations, and their specific application to fully connected and convolutional layers.
The compression method for convolutional layers is a key novelty, as few previous papers  extend factorization-based compression methods beyond fully connected layers.
{Finally, we show our experimental results improve upon the state-of-the-art in compressibility without significant performance degradation } in Section~\ref{sec:exp} and conclude with future work in Section~\ref{sec:conclude}


\section{Related Work}
\label{sec:relWork}
Past deep neural network compression techniques have largely applied to fully connected layers, which previously have dominated the number of parameters of a model. However, since modern models like ResNet and WideResNet are moving toward wider convolutional layers and omitting fully connected layers altogether, it is important to consider compression schemes that work on both fronts. 

Many modern compression schemes focus on post-processing techniques, such as hashing \cite{chen2015compressing} and quantization \cite{gong2014compressing}. A strength of these methods is that they can be applied in addition to any other compression scheme, and are thus orthogonal to other methods.
More similar to our work are novel representations like circulant projections \cite{cheng2015fast} and truncated SVD representations \cite{denton2014exploiting}. 

Low-rank tensor approximation of deep neural networks has been widely investigated in the literature for effective model compression, low generative error, and fast prediction speed \cite{sokolic2017generalization, kim2015compression, lebedev2014speeding}.
Tensor Networks (TNs) \cite{cichocki2016tensor,cichocki2017tensor} 
have recently drawn considerable attention in multi-dimensional data representation \cite{wang2016tensor,wang2017efficient,dai2006tensor,wang2017tensor}, and deep learning \cite{cohen2016expressive,cohen2016convolutional,cohen2016deep,kossaifi2017tensor}. 

One of the most popular methods of tensor factorization is the Tucker factorization \cite{tucker1966some}, 
and has been shown to exhibit  good performance in data representation \cite{dai2006tensor,ashraphijuo2017characterization,ashraphijuo2016deterministic}  and in compressing  fully connected layers in deep neural networks \cite{kossaifi2017tensor}. 
In \cite{kim2015compression}, a Tucker decomposition approach is applied to compress both fully connected layers and convolution layers.

Tensor train (TT) representation  \cite{oseledets2011tensor} is another example of TNs that factorizes a tensor into {boundary two matrices and a set of $3^\text{rd}$ order tensors}, 
and has demonstrated its capability in data representation \cite{phien2016efficient,wang2016tensor,ashraphijuo2017rank} and deep learning \cite{novikov2015tensorizing,yang2017tensor}.
In \cite{wang2017efficient}, the TT model is compared against TR for multi-dimensional data completion, showing that for the same intermediate rank, TR can be far more expressive than TT, motivating the generalization. 
In this paper, we investigate TR for deep neural network compression.

\section{Tensor Ring Nets (TRN)}
\label{sec:model}

In this paper,  $\mathscr{X} \in \mathbb{R}^{I_1 \times \cdots \times I_d}$ is a $d$ mode tensor with $\prod_{i=1}^d I_i$ degrees of freedom.
A \emph{tensor ring decomposition} factors such an $\mathscr X$  into $d$ independent $3$-mode tensors, $\mathscr{U}^{(1)},\ldots,\mathscr{U}^{(d)}$
such that each entry inside the tensor $\mathscr{X}$ is represented as
\begin{equation}
\mathscr{X}_{i_1, \cdots, i_d} = \sum_{r_1,\cdots,r_d}\mathscr{U}^{(1)}_{r_d, i_1, r_1}\mathscr{U}^{(2)}_{r_1, i_2, r_2} \cdots \mathscr{U}^{(d)}_{r_{d-1}, i_d, r_d},
\label{eq:tensor_net_decomp}
\end{equation}
where $\mathscr U^{(i)} \in \mathbb{R}^{R \times I_i \times R}$, and  $R$ is the \emph{tensor ring rank}. 
\footnote{More generally, $\U^{(i)} \in \mathbb{R}^{R_i \times I_i \times R_{i+1}}$ and each $R_i$ may not be the same. For simplicity, we assume $R_1 = \cdots = R_d = R$.
}
Under this low-rank factorization, the number of free parameters is reduced to $R^2\sum_{i=1}^{d} I_i$ in the tensor ring factor form,  which is significantly less than $\prod_{i=1}^d I_i$ in $\mathcal{X}$.

For notational ease, let $\mathscr{U} = \{\mathscr{U}^{(1)}, \cdots, \mathscr{U}^{(d)}\}$, and define $\decomp(\mathscr X;R,d)$ as the operation to obtain $d$ factors $\mathscr{U}^{(i)}$ with tensor ring rank $R$ from $\mathscr{X}$, and $\construct(\mathscr U)$ as the operation to obtain $\mathscr{X}$ from $\mathscr{U}$. 
	
Additionally, for $1\leq k<j \leq d$, define the {\bf merge} operation as $\mathscr{M} =\merge ({\mathscr U}, k,j)$ such that $\mathscr{U}_k,\mathscr{U}_{k+1},\cdots, \mathscr{U}_j$ are merged into one single tensor $\mathscr{M}$ of dimension $R \times I_k\times I_{k+1}\times \cdots \times I_j \times R$, and each entry in $\mathscr{M}$ is
	\begin{equation}
	\begin{split}
	 &\mathscr{M}_{r_{k-1}, i_k,i_{k+1}, \cdots, i_j, r_j}=
	 \\ 
	 &\sum_{r_k,\cdots,r_{j-1}}\mathscr{U}^{(k)}_{r_{k-1}, i_k, r_k}\mathscr{U}^{(k+1)}_{r_k, i_{k+1}, r_{k+1}} \cdots \mathscr{U}^{(j)}_{r_{j-1}, i_j, r_j}.
	\end{split}
	\end{equation}
	Note that construct operator is the merge operation $\merge(\mathscr{U}, 1,d)$, which results in a tensor of shape $R \times I_1 \times I_2 \times \cdots \times I_d \times R$, followed by summing along mode $1$ and mode $d+2$, resulting in a tensor of shape $I_1 \times I_2 \times \cdots \times I_d$; e.g.
	\[
	\construct(\U) = \sum_{r=1}^R\merge(\U, 1, d)_{r, :,r}.
 	\]
	
\paragraph{Tensor diagrams}
\begin{figure}
	\begin{center}
	\includegraphics[width=3in]{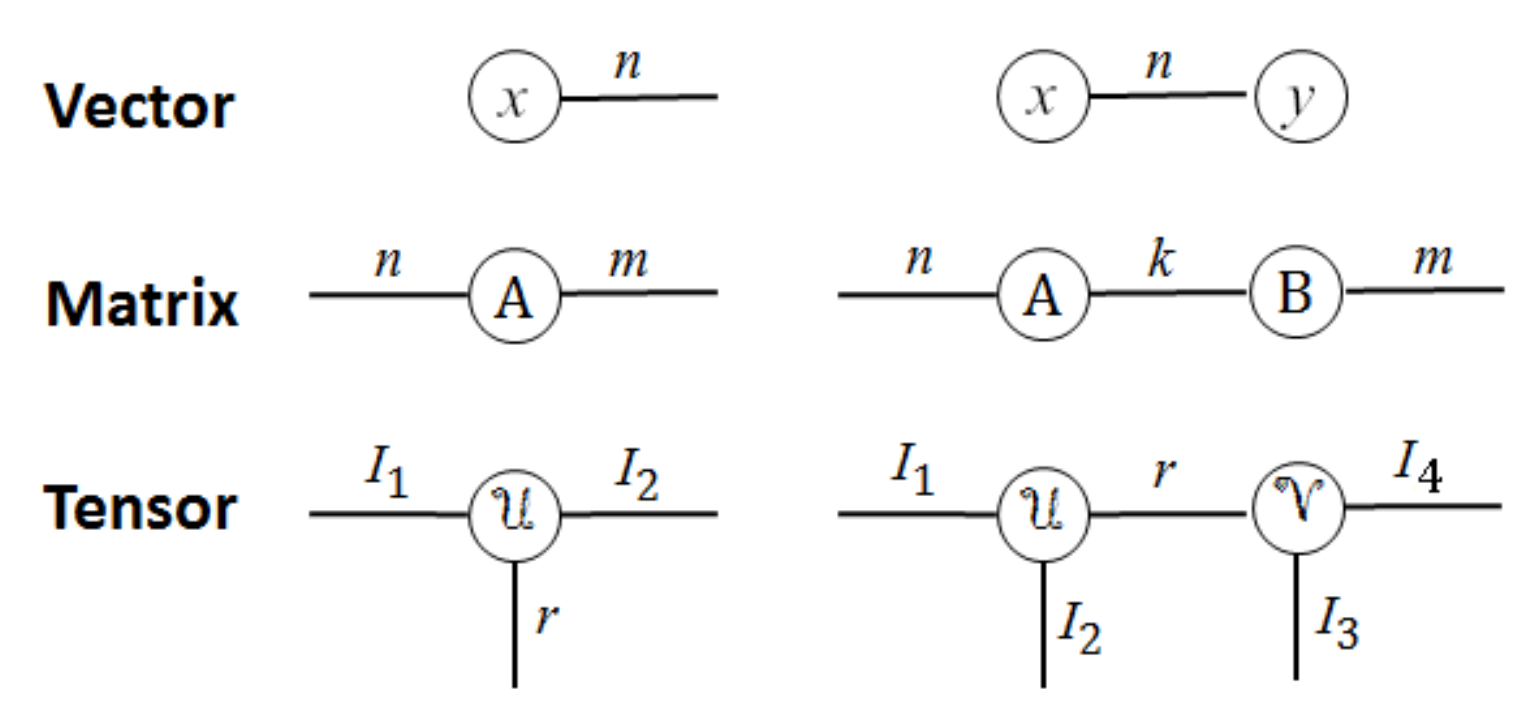}
	\caption{\textbf{Tensor diagrams.} Left: A graphical representation of a length $n$ vector $x$, a $n\times m$ matrix $A$, and a 3rd order $I_1\times I_2 \times I_3$ tensor $\U$. Right: factorized forms for a dot product $x^Ty$, matrix product $AB$ where $A$ and $B$ have $k$ rows and columns respectively, and the tensor product of  $\U$ and $\mathscr V$ along a common axis. More explicitly, the tensor product on the bottom right has 4 orders and the $i_1,i_2,i_3,i_4$-th element is $\sum_{j=1}^r\U_{i_1,i_2,j}\mathscr V_{i_3,i_4,j}$ for $i_k = 1,\ldots, I_k$, $k = 1,2,3,4$.}
	\vspace{-.3in}
	\label{f-tensor}
	\end{center}
\end{figure}

\begin{figure}
	\begin{subfigure}[t]{0.22\textwidth}
	\includegraphics[width=1.5in]{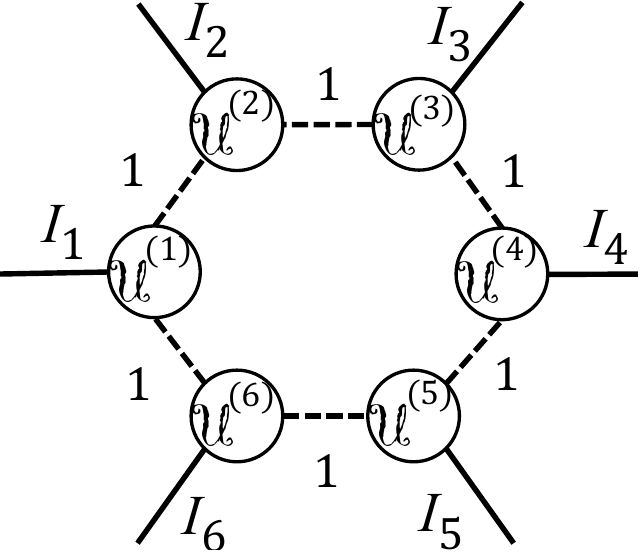}
	\caption{CP Decomposition}
	\end{subfigure}
\hspace{.1in}
\begin{subfigure}[t]{0.22\textwidth}
	\includegraphics[width=1.5in]{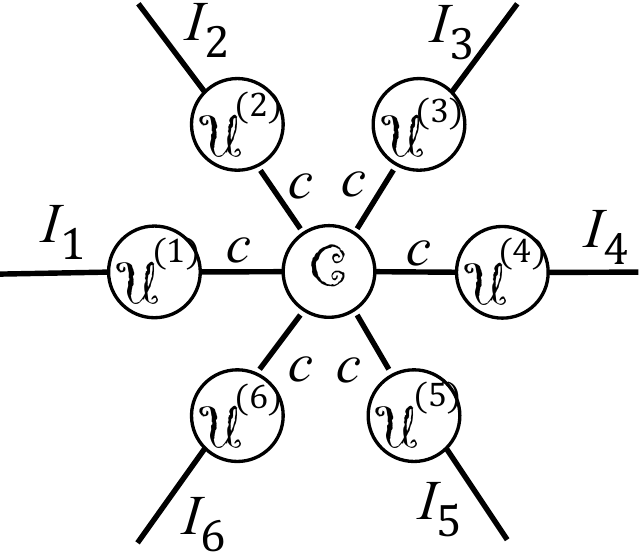}
	\caption{Tucker}
\end{subfigure}
	\begin{subfigure}[t]{0.22\textwidth}
	\includegraphics[width=1.5in]{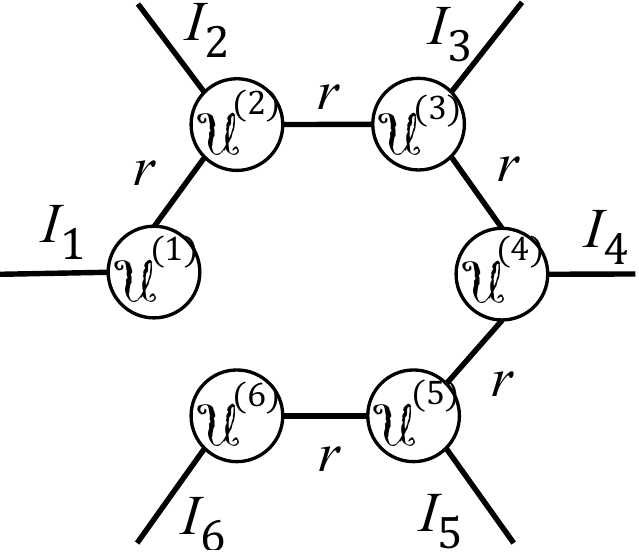}
	\caption{Tensor Train (TT)}
	\end{subfigure}
\hspace{.1in}
	\begin{subfigure}[t]{0.22\textwidth}
	\includegraphics[width=1.5in]{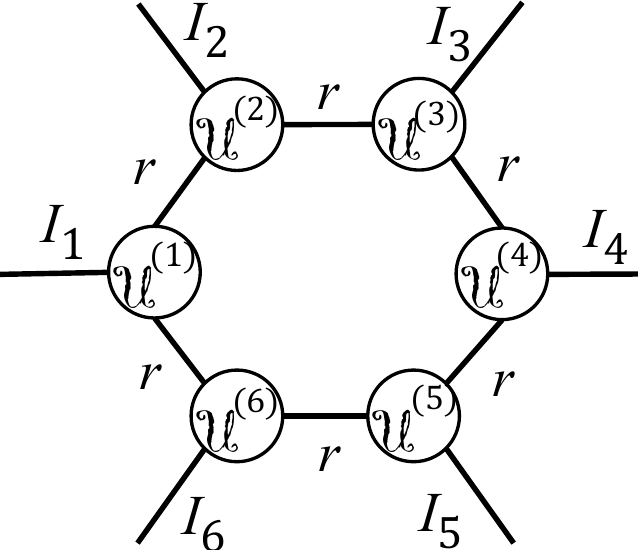}
	\caption{Tensor Ring(TR)}
	\end{subfigure}
	\caption{\textbf{Tensor decompositions.} Tensor diagrams for four popular tensor factorization methods: (a) the CP decomposition (unnormalized), (b) the Tucker decomposition, (c) the Tensor Train (TT) decomposition, and (d) the Tensor Ring (TR) decomposition used in this paper. 
	As shown, TR can be viewed as a generalization of both CP (with $r > 1$) and TT (with an added edge connecting the first and last tensors). 
	In Section \ref{sec:exp}, we also compare against Tucker decomposition compression schemes.}
	\label{f-tensor-ring}
	\vspace{-0in}
\end{figure}

Figure \ref{f-tensor} introduces the popular tensor diagram notation \cite{orus2014practical}, which represents tensor objects as nodes and their axes as edges of an undirected graph. An edge connecting two nodes indicates multiplication along that axis, and a ``dangling" edge shows an axis in the remaining product, with the dimension given as the edge weight. This compact notation is  useful in representing various factorization methods (Figure \ref{f-tensor-ring}).

\vspace{-.1in}
\paragraph{Merge ordering}
The computation complexity in this paper is measured in flops (counting additions and multiplications).
The number of flops for a \construct~  depends on the sequence of merging $\mathscr{U}^{(i)}, i=1,\cdots,d$. (See figure \ref{f-merge-order}).
A detailed analysis of the two schemes is given in appendix \ref{app:merge}, resulting in the following conclusions.
\begin{theorem}
	\label{th-mergeordering}
	Suppose $I_1 = \cdots = I_d\geq 2$ and $I = \prod_{i=1}^d I_i$. Then 
	\begin{enumerate}
		\vspace{-.1in}
		\item any merge order costs between $2R^{3}I$ and $4R^3I$ flops,
		\vspace{-.1in}
		\item any merge order costs requires storing between $R^2 I$ and $2R^2I$ floats, and
		\vspace{-.1in}
		\item if $d$ is a power of 2, then a hierarchical merge order  achieves the minimum flop count.
	\end{enumerate}
\end{theorem}
\vspace{-.2in}
\begin{proof}
	See appendix \ref{app:merge}.
\end{proof}

\begin{figure}
	\begin{center}
		\includegraphics[width=3.25in]{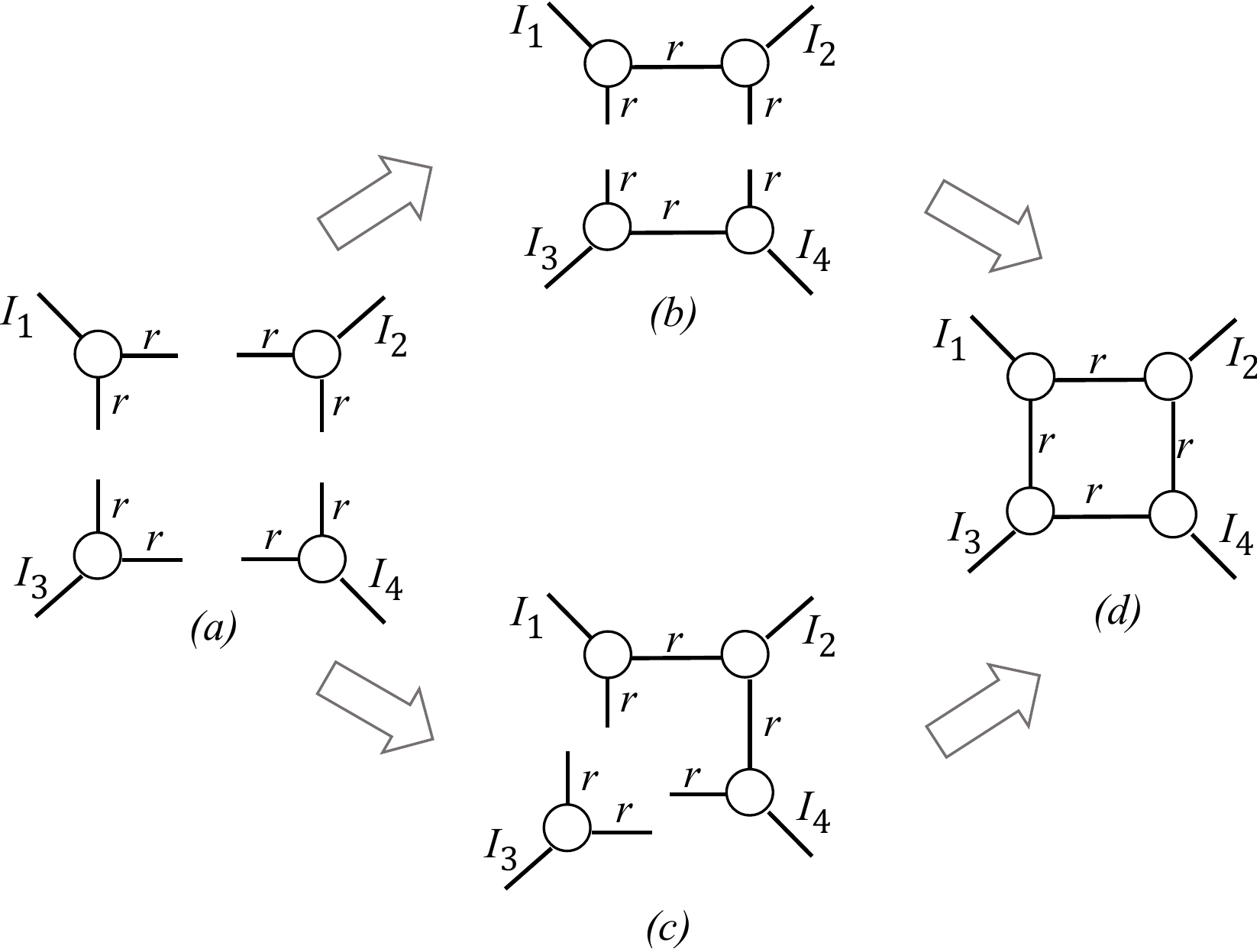}
		\caption{\textbf{Merge ordering.} A 4th order tensor is merged from its factored form, either hierarchically via (a)$\to$(b)$\to$(d), or sequentially via (a)$\to$(c)$\to$(d). Note that the computational complexity of forming (b) is {$r^3 (I_1I_2+I_3I_4)$} and for (c) is {$r^3 (I_1I_2 + I_1I_2I_4)$}, and (c) is generally more expensive (if $I_1\approx I_2 \approx I_3 \approx I_4$). This is discussed in detail in Appendix \ref{app:merge}.}
		\label{f-merge-order}
		\vspace{-.2in}
	\end{center}
\end{figure}

Several interpretations can be made from these observations. First, though different merge orderings give different flop counts, the worst choice is at most 2x more expensive than the best choice. 
However, since we have to make some kind of choice, we note that since every merge order is a combination of hierarchical and sequential merges, striving toward a hierarchical merging is a good heuristic to minimize flop count.
Thus, in our paper, we always use this strategy.

A \emph{Tensor Ring Network (TRN)} is a tensor factorization of either fully connected layers (FCL) or convolutional layers (ConvL), trained via back propagation. 
If a pre-trained model is given, a good initialization can be obtained from the tensor ring decomposition of the layers in the pre-trained model.

\subsection{Fully Connected Layer Compression}
\begin{figure}
	\begin{center}
		\includegraphics[width=3in]{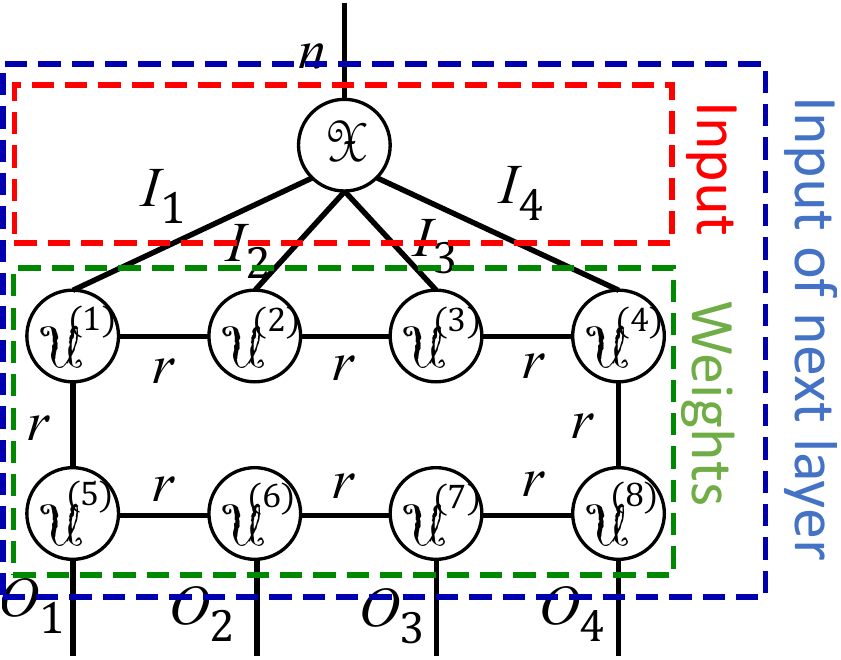}
		\caption{\textbf{Fully connected layer.} Tensor diagram of a fully connected TRN, divided into input and weights. The composite tensor is the input into the next layer.}
		\vspace{-.2in}
		\label{f-fully}
	\end{center}
\end{figure}

In feed-forward neural networks, an input feature vector ${\bf x}\in \mathbb{R}^{I}$ is mapped to an output feature vector ${\bf y} = {\bf Ax}\in \mathbb{R}^{O}$ via a fully connected layer ${\bf A}\in \mathbb{R}^{I \times O}$. 
Without loss of generality, $\bf{x}$, $\bf{A}$, and $\bf y$ can be reshaped into higher order  tensors $\mathscr{X}$, $\mathscr{A}$, and $\mathscr {Y}$ with 
\begin{equation}\label{eq: FC}
\mathscr{Y}_{o_1,\ldots, o_{\hat d}} = \sum_{i_1,\ldots,i_{d}}
 \mathcal{A}_{i_1,\ldots, i_{d},o_1,\ldots, o_{\hat d}} \mathcal{X}_{i_1,\ldots, i_{d}}
\end{equation}
where $d$ and $\hat d$ are the modes of $\mathscr{X}$ and $\mathscr{Y}$ respectively, and 
$i_k$'s ad $o_k$'s span from $1$ to $I_k$ and 1 to $O_k$ respectively, and 
\[
\prod_{i=1}^d I_i = I, \quad \prod_{i=1}^{\hat d} O_i = O.
\]
To compress a feed-forward network, we decompose as $\mathscr{U} =\{
\mathcal{U}^{(1)}, \ldots, \mathcal{U}^{(d+{\hat d})}
\}
= \decomp( \mathcal{A}; R, d+{\hat d} )$
and replace $\mathscr A$ with its decomposed version in \eqref{eq: FC}.
A tensor diagram for this operation is given in Figure \ref{f-fully}, which shows how each multiplication is applied and the resulting dimensions.

\paragraph{Computational cost}
The computational cost again depends on the order of merging $\mathscr{X}$ and $\mathscr{U}$.
{Note that there is no need to fully construct the tensor $\mathscr A$, and a tensor representation of $\mathscr A$ is sufficient to obtain $\mathscr Y$ from $\mathscr X$.}
To reduce the computational cost, a \emph{layer separation} approach is proposed by first using hierarchical merging to obtain 
\begin{equation}
\begin{split}
\mathscr{F}^{(1)} &= \merge(\mathscr{U}, 1, d) \in \mathbb{R}^{R \times I_1 \times \cdots \times I_d \times R}\\
\mathscr{F}^{(2)} &=\merge(\mathscr{U}, d+1, d+\hat{d}) \in
\mathbb{R}^{R \times O_{1} \times \cdots \times O_{\hat{d}} \times R},
\end{split}
\end{equation}
which {is upper bounded by $4R^3 (I+O)$} flops. 
By replacing $\mathscr{A}$ in \eqref{eq: FC} with  $\mathscr{F}^{(1)}$ and $\mathscr{F}^{(2)}$ and switching the order of summation, we obtain
\begin{eqnarray}
\mathscr{Z}_{r_d,r_{d+\hat d}}&=& 
\sum_{i_1,\ldots,i_{d}}\mathscr{F}^{(1)}_{r_{d+\hat{d}}, i_1,\cdots, i_d, r_d}
\mathcal{X}_{i_1,\ldots, i_{d}},
\label{eq: FC_Sep-a}\\
\mathscr{Y}_{o_1,\ldots, o_{\hat d}}&=&
\sum_{r_{d+\hat{d}}, r_d}
\mathscr Z_{r_d,r_{d+\hat d}}
\mathscr{F}^{(2)}_{r_d, o_1,\cdots, o_{\hat d}, r_{d+\hat{d}}}.
\label{eq: FC_Sep}
\end{eqnarray}
The summation \eqref{eq: FC_Sep-a} 
is equivalent to a feed-forward layer of shape $(I_1\cdots I_d) \times R^2$, which takes $2R^2 I$ flops. 
Additionally, the summation over $r_{d+\hat{d}}$ and $r_d$ is equivalent to another feed-forward layer of shape $R^2 \times (O_{1}\cdots O_{\hat{d}})$, which takes $2R^2O$ flops. 
Such analysis demonstrates that the \emph{layer separation} approach to a FCL in a tensor ring net is equivalent to a low-rank matrix factorization to a fully-connected layer, thus reducing the computational complexity {when $R$ is relatively smaller than $I$ and $O$}.

Define $P_{\text{FC}}$ and $C_{\text{FC}}$ as the complexity saving in parameters and computation, respectively, for the tensor net decomposition over the typical fully connected layer forward propagation. Thus we have
\begin{equation}
P_{\text{FC}} = 
\frac{IO }{R^2\left(\sum_i^d I_i +\sum_j^{\hat d} O_j\right)}.
\end{equation}
and
\begin{equation}
\label{eq:fully-C}
C_{\text{FC}}\geq 
\frac{2BIO }
{(4R^3+2 BR^2)(I+O)},
\end{equation}
where $B$ is the batch size of testing samples. Here, we see the compression benefit in computation; when $B$ is very large, \eqref{eq:fully-C} converges to $IO/(R^2(I+O))$, which for large $I$, $O$ and small $R$ is significant. Additionally, though the expensive reshaping step grows cubically with $R$ (as before), it does not grow with batch size; conversely, the multiplication itself (which grows linearly with batch size) is only quadratic in $R$. 
In the paper, the parameter is selected by picking small $R$ and large $d$ to achieve the optimal {$C$ since $R$ needs to be small enough for computation saving}.
\
\label{sec:fully}

\subsection{Convolutional Layer Compression}
In convolutional neural networks(CNNs), an input tensor $\mathcal{X} \in \mathbb{R}^{H\times W \times I}$ is 
convoluted with a $4$th order kernel tensor $\mathcal{K}\in \mathbb{R}^{D \times D \times I \times O}$ and mapped to a $3$rd order tensor $\mathcal{Y}\in \mathbb{R}^{H\times W \times O}$, as follows
\begin{equation}\label{eq: Conv}
\begin{split}
	\mathcal{Y}_{h, w, o} &= \sum_{d_1, d_2=1}^{D}\sum_{i=1}^{I}
	\mathcal{X}_{h', w',i}
	\mathcal{K}_{d_1,d_2, i, o},
	\\
	h' &=(h-1)s + d_1-p,\\
	w'&=(w-1)s+d_2-p,
\end{split}
\end{equation}
where $s$ is stride size, $p$ is zero-padding size.
Computed as in \eqref{eq: Conv}, the flop cost is $D^{2}\cdot IO\cdot HW$. \footnote{For small filter sizes $D \ll \log(HW)$, as is often the case in deep neural networks for image processing, often direct multiplication to compute convolution is more efficient than using an FFT, which for this problem has order $IO(HW(\log(HW)))$ flops. Therefore we only consider direct multiplication as a baseline.}

\begin{figure}
		\begin{center}
			\includegraphics[width=2.25in]{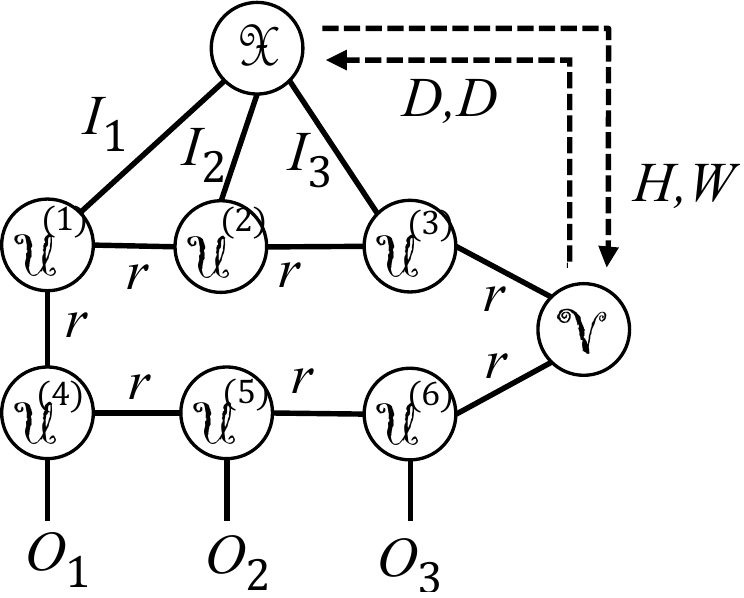}
			\caption{\textbf{Convolutional layer.} Dashed lines show the convolution operation \eqref{eq: Conv}. Here, $\U^{(1)}$, $\U^{(2)}$ and $\U^{(3)}$ decompose $\U$ and $\U^{(4)}$, $\U^{(5)}$, and $\U^{(6)}$ decompose $\hat \U$ in  \eqref{eq: TR_Kernel}. The dashed line between $\mathscr X$ and $\V$ represent the convolution operation as expressed in \eqref{eq: Conv}. Note that $I_1\times I_2 \times I_3$ decompose the number of \emph{channels} entering the layer (which is 1 at the first input), where in Figure \ref{f-fully} they decompose the \emph{feature dimension} entering the layer.}
			\label{f-convolution}
		\end{center}
		\vspace{-.3in}
	\end{figure}

In TRN, tensor ring decomposition is applied onto the kernel tensor $\mathcal{K}$ and factorizes the $4$th order tensor into four $3$rd tensors.
With the purpose to maintain the spatial information in the kernel tensor, we do not factorize the spatial dimension of $\mathcal{K}$ via merging the spatial dimension into one $4th$ order tensor $\V^{(1)}_{R_1, D_1, D_2, R_2}$, thus we have
\begin{equation}\label{eq: TR_Kernel}
\begin{split}
\mathcal{K}_{d_1, d_2, i, o}  = \sum_{r_1, r_2, r_3=1}^{R}
\mathcal{V}_{r_1, d_1, d_2, r_2}
\mathcal{U}_{r_2, i, r_3}
\mathcal{\hat U}_{r_3, o, r_1}.
\end{split}
\end{equation}
{In the scenario when $I$ and $O$ are large}, the tensors $\U$ and $\hat \U$ are further decomposed into $\U^{(1)},\ldots,\U^{(d)}$ and $\U^{(d+1)},\ldots,\U^{(d+\hat d)}$ respectively.
(See also Figure \ref{f-convolution}.)

The kernel tensor factorization in \eqref{eq: TR_Kernel} combined with the convolution operation in \eqref{eq: Conv} can be equivalently solved in three steps:
\begin{eqnarray}
\mathscr{P}_{h', w', r_2, r_3} &=& \sum_{i=1}^I
\mathscr{X}_{h',w',i}\mathscr{U}^{(2)}_{r_2, i, r_3}\label{eq:conv-1}\\
\mathscr{Q}_{h, w, r_3, r_1} &=& \sum_{d_1,d_2=1}^D \sum_{r_2}^R \mathscr{P}_{h',w', r_2, r_3} \mathscr{U}^{(1)}_{r_1,d_1,d_2,r_2}\label{eq:conv-2}\\
\mathscr{Z}_{h,w,o} &=& \sum_{r_1, r_3}
\mathscr{Q}_{h, w, r_3, r_1}
\mathscr{U}^{(3)}_{r_3, o, r_1}.\label{eq:conv-3}
\end{eqnarray}
where \eqref{eq:conv-1} is a tensor multiplication along one slice, with flop count $HWR^2I$, \eqref{eq:conv-2} is a 2-D convolution with flop count $HWR^3D^2$, and \eqref{eq:conv-3} is a tensor multiplication along 3 slices with flop count $HWR^2O$.
{This is also equivalent to a three-layer convolutional networks without non-linear transformations, where 
\eqref{eq:conv-1} is a convolutional layer from $I$ feature maps to $R^2$ feature maps with a $1\times 1$ patch,
\eqref{eq:conv-2} contains $R$ convolutional layers from $R$ feature maps to $R$ feature maps with a $D\times D$ patch,
and \eqref{eq:conv-3} is a convolutional layer from $R^2$ feature maps to $O$ feature maps with with a $1 \times 1$ patch.
This is a common sub-architecture choice in other deep CNNs, like the inception module in GoogleNets~\cite{szegedy2015going}, but without nonlinearities between $1\times1$ and $D\times D$ convolution layers.}


{\bf Complexity:} We employ the ratio between complexity in CNN layer and the complexity in tensor ring layer to quantify the capability of TRN in reducing computation ($C_{\text{conv}}$) and parameter ($P_{\text{conv}}$) costs, 
\begin{equation}\label{eq: MC1}
\begin{split}
&P_{\text{conv}} =\frac{D^2IO}{D^2R^2 + IR^2 + OR^2},\\
&C_{\text{conv}} = \frac{IO \cdot D^2}{R^2I + R^3D^2 + R^2O}.
\end{split}
\end{equation}
If, additionally, the tensors $\U^{(1)}$ and $\U^{(2)}$ are further decomposed to $d$ and $\hat d$ tensors, respectively, then
\begin{equation}\label{eq: MCd}
\begin{split}
&P_{\text{conv}} =\frac{D^2IO}{D^2R^2 + R^2(\sum_i^d I_i + \sum_j^{\hat d} O_j)},\\
&C_{\text{conv}} = \frac{BIO \cdot D^2}{4R^3{(I+O)} + BR^2(I+O) + BR^3D^2 }.
\end{split}
\end{equation}
Note that in the second scenario, we have a further compression in storage requirements, but lose gain in computational complexity, which is a design tradeoff.
{In our experiments, we further factorize $\U^{(1)}$ and $\U^{(3)}$ in to higher order tensors in order to achieve our gain in model compression.}

\label{sec:conv}

\vspace{-.15in}
\paragraph{Initialization}
In general nonconvex optimization (and especially for deep learning) the choice of initial variables can dramatically effect the quality of the model training. In particular, we have found that initializing each parameter randomly from a Gaussian distribution is effective, with a carefully chosen variance. 
If we initialize all tensor factors as drawn i.i.d. from $\mathcal N(0,\sigma^2)$, then after merging $d$ factors the merged tensor elements will have mean 0 and variance $R^d\sigma^{2d}$ (See appendix \ref{ini_proof}).
By picking $\sigma = \left(\frac{2}{N}\right)^{1/d}\frac{1}{\sqrt{R}}$, where $N$ is the amount of parameters in the uncompressed layer, the merged tensor will have mean 0, variance $\sqrt{2/N}$, and in the limit will also be Gaussian. Since this latter distribution  works well in  training the uncompressed models, choosing this value of $\sigma$ for initialization is well-motivated, and observed to be necessary for good convergence.

\label{sec:init}

\section{Experiments}

We now evaluate the effectiveness of TRN-based compression on several well-studied deep neural networks and datasets: LeNet-300-100 and LeNet-5 on MNIST, and ResNet and WideResNet on Cifar10 and Cifar100.
These networks are trained using Tensorflow~\cite{abadi2016tensorflow}. {All the experiments on LeNet are implemented on Nvidia GTX 1070 GPUs, and all the experiments for ResNet and WideResNet are implemented on Nvidia GTX Titan X GPUs.}
In all cases, the same tensor ring rank $r$ is used in the networks, and all the networks are trained from randomly initialization using the the proposed initialization method.
Overall, we show that this compression scheme can give significant compression gains for small accuracy loss, and even negligible compression gains for no accuracy loss.

\subsection{Fully connected layer compression}
The goal of compressing the LeNet-300-100 network is to assess the effectiveness of compressing fully connected layers using TRNs; as the name suggests, LeNet-300-100 contains two hidden fully connected layers with output dimension 300 and 100, and an output layer with dimension 10 (= \# classes). 
Table \ref{ana_fc_table} gives the parameter settings for LeNet-300-100, both in its original form (uncompressed) and in its tensor factored form. 
A compression rate greater than 1 is achieved for all {$r \leq 54$}, and a reduction in computational complexity for all $r \leq 6$; both are typical choices.

Table \ref{res_fc_table} shows the performance results on MNIST classification for the original model (as reported in their paper), and compressed models using both matrix factorization and TRNs. For a 0.14\% accuracy loss, TRN can compress up to $13\times$, and for no accuracy loss, can compress $1.2\times$. Note also that matrix factorization, at $16\times$ compression, performs worse than TRN at $117\times$ compression, suggesting that the high order structure is helpful.  
{Note also that low rank Tucker approximation in \cite{kim2015compression} is equivalent to low rank matrix approximation when compressing fully connected layer. }



\begin{table*}[t!]
	\begin{center}
		\begin{tabular}{ |c|c|c|c|c|c|c| }
			\hline
			&\multicolumn{3}{c|}{Uncompressed dims.} & \multicolumn{3}{c|}{TRN dimensions}\\
			\hline
			layer 		&shape 				& \# params 			&flops 		&shape of composite tensor &  \# params &flops\\
			\hline
			fc1			&$784 \times 300$ 		& 235K 			&470K		&$(4 \times 7 \times 4 \times 7) \times (3 \times 4 \times 5 \times 5)$ 	&$39r^2$		&$1177r^3+1084r^2$\\
			fc2 			&$300\times 100$ 		& 30K 			&60K		&$(3 \times 4 \times 5 \times 5) \times (4 \times 5 \times 5)$ 			&$31r^2$ 		&$457r^3 + 400r^2$ \\
			fc3 			&$100 \times 10$		& 1K				&2K		&$(4\times 5 \times 5) \times (2 \times 5)$						&$21r^2$ 		&$127r^3 + 107r^2$ \\
			\hline
			Total 			&-					& 266K			&532K		&-													&$91r^2$		&$1761r^3 + 1591r^2$\\
			\hline
		\end{tabular}
	\end{center}
	\vspace{-.15in}
	\caption{\textbf{Fully connected compression.} Dimensions of the three-fully-connected layers in the uncompressed (left) and TRN-compressed (right) models.  The computational complexity includes  tensor product merging ($O(r^3)$)  and feed-froward multiplication ($O(r^2)$).}
	\label{ana_fc_table}
	\vspace{-.1in}
\end{table*}

\begin{table*}[t!]
	\begin{center}
\begin{tabular}{ |l|c|c|c|c|c|c| }
\hline
Method 							&Params 		& CR	&Err \% 		&Test (s) 	& Train (s/epoch) 	&LR\\
\hline
LeNet-300-100 \cite{lecun1998gradient}	&266K 		& $1	  \times$ 			&$2.50$		&$0.011\pm0.002$ 	& $3.5\pm1.0$ 			&$2e^{-4}$\\
\hline
M-FC\cite{denton2014exploiting,kim2015compression}($r=10$)	&16.4K 		& $16.3 \times$			&$3.91$		& $0.016 \pm 0.010$ 			& $6.4 \pm 1.2$			&$1e^{-4}$\\
M-FC ($r=20$)										&31.2K 		&$5.3\times$ 			&$3.0$		& $0.014 \pm 0.010$ 			& $5.2 \pm 1.2$			&$1e^{-4}$\\
M-FC ($r=50$)										&75.7K 		&$3.5\times$			&$2.62$		& $0.021 \pm 0.012$ 			& $8.1 \pm 1.2$			&$1e^{-4}$\\
\hline
TRN ($r=3$)							&0.8K 		& $325.5 \times$		&$8.53$		&$0.015\pm0.007$ 	& $7.9\pm1.4$			&$1e^{-3}$\\
TRN ($r=5$)	 						&2.3K 		& $117.2 \times$		&$3.75$		&$0.015\pm0.007$ 	& $7.8\pm1.4$			&$2e^{-3}$\\
TRN ($r=15$)							&20.5K 		& $13.0\times$ 			&$2.64$ 		&$0.015\pm0.007$ 	& $8.1\pm1.4$			&$5e^{-4}$\\
TRN ($r=50$)							&227.5K 		& $1.2 \times$			&${\bf 2.31}$	&$0.022\pm0.008$ 	& $11.1\pm1.4$			&$5e^{-5}$\\
\hline
\end{tabular}
\end{center}
\vspace{-.15in}
\caption{\textbf{Fully connected results.}  LeNet-300-100 on MNIST datase, trained to 40 epochs, using a minibatch size 50. Trained from random weight initialization.   
{ADAM \cite{kingma2014adam} is used for optimization. Testing time is per 10000 samples.} CR = Compression ratio. LR = Learning rate.}
\label{res_fc_table}
\end{table*}

\subsection{Convolutional layer compression}
We now investigate compression of convolutional layers in a small network. 
LeNet-5 is a (relatively small) convolutional neural networks with 2 convolution layers, followed by 2 fully connected layers, which achieves $0.79\%$ error rate on MNIST. 
The dimensions before and after compression are given in Table \ref{lenet5_ana}.
 In this wider network we see a much greater potential for compression, with positive compression rate whenever  $r \leq 57$. However, the reduction in complexity is more limited, and only occurs when
 $r\leq 4$. 
 
 However, the performance on this experiment is still positive. 
 By setting $r=20$, we compress LeNet-5 by $11\times$ and a lower error rate than the original model as well as the  Tucker factorization approach. If we also require a reduction in flop count, we incur an error of 2.24\%, which is still quite reasonable in many real applications.

\begin{table*}[t!]
	\begin{center}
	\begin{tabular}{ |c|c|c|c|c|c|c| }
		\hline
		&\multicolumn{3}{c|}{Uncompressed dims.} & \multicolumn{3}{c|}{TRN dimensions}\\
		\hline
		layer 		&shape 						& \# params  			&flops 		&shape 									& \# params 		&flops\\
		\hline
		conv1 		&$5\times 5 \times 1 \times 20$		& 0.5K			&784K		&$5\times 5 \times 1\times (4 \times 5)$						&$19r^2$ 			&$33408r^2 + 39245r^3$\\
		conv2 		&$5 \times 5 \times 20 \times 50$	& 25K			&5000K		&$5 \times 5 \times (4 \times 5) \times (5 \times 10)$ 		&$34r^2$	 		&$17840r^2 + 5095r^3$\\
		fc1 			&$1250 \times 320$				& 400K			&800K		&$(5 \times 5 \times 5 \times 10) \times (5 \times 8 \times 8)$	&$46r^2$			&$1570r^2   + 1685r^3$\\
		fc2			&$320 \times 10$				& 3K				&6K			&$(5 \times 8 \times 8) \times 10$					&$31r^2$ 			&$330r^2	   + 360r^3$\\
		\hline
		Total 			&-							& 429K			&6590K	 	&- 											&$130r^2$ 		&$53148r^2 + 46385r^3$\\			
		\hline
	\end{tabular}
\end{center}
\vspace{-.2in}
	\caption{\textbf{Small convolution compression.} Dimensions of LeNet-5 layers in its original form (left) and TRN-compressed (right). 
	The computational complexity includes tensor product merging and convolution operation in \eqref{eq:conv-2} of $O(r^3)$,  and convolution in \eqref{eq:conv-1} \eqref{eq:conv-3} of $O(r^2)$.
	}
	\label{lenet5_ana}
\end{table*}

\begin{table*}[t!]
	\begin{center}
\begin{tabular}{ |l|c|c|c|c|c|c| }
\hline
Method 						&Params 		& CR	&Err \% 		&Test (s) 	& Train (s/epoch) 	&LR\\
\hline
LeNet-5 \cite{lecun1998gradient}	&429K 		& $1	  \times$ 			&$0.79$		&$0.038 \pm 0.027$ 			&$1.6\pm1.9$ 		&$5e^{-4}$\\
\hline
Tucker \cite{kim2015compression}	&189K 		& $2\times$			&$0.85$ 		&$0.066 \pm 0.025$ 			&$7.7\pm 3$ 		&$5e^{-4}$ \\
\hline
TRN ($r=3$)						&1.5K 		& $286\times$ 			&$2.24$ 		&$0.058 \pm 0.026 $  		&$8.3\pm4.5$ 		&$5e^{-4}$\\
TRN ($r=5$)						&3.6K 		& $120\times$			&$1.64$		&$0.072 \pm0.039$  			&$10.6\pm7.1$ 		&$5e^{-4}$\\
TRN ($r=10$)					&11.0K 		& $39\times$ 			&$1.39$		&$0.080 \pm 0.025$  		& $15.6\pm4.6$		&$2e^{-4}$\\
TRN ($r=15$)					&23.4K		& $18\times$			&$0.81 $		& $0.039\pm 0.019$ 			&$20.1\pm16.0$ 		&$2e^{-4}$\\
TRN ($r=20$)					&40.7K		& $11\times$ 			&${\bf 0.69}$	&$0.052\pm 0.028$  			&$27.8\pm7.4$ 		&$1e^{-5}$\\
\hline
\end{tabular}
\end{center}
\vspace{-.2in}
\caption{\textbf{Small convolution results.}  LeNet-5 on MNIST dataset, trained to 20 epochs, using a minibatch size 128.   {ADAM \cite{kingma2014adam} is used for optimization. Testing time is per 10000 samples}. CR = Compression ratio. LR = Learning rate.}
\end{table*}

\begin{figure}[t!]
	\includegraphics[width=0.48\textwidth]{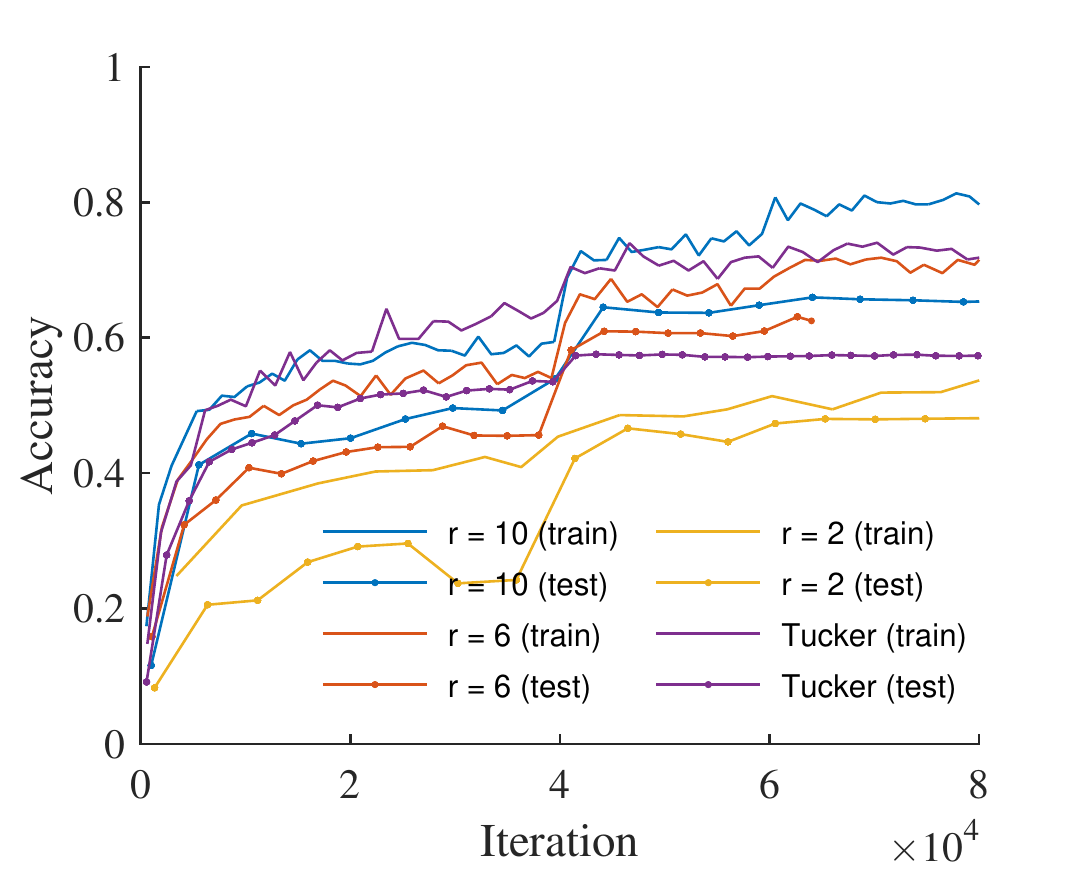}
	\centering
	\vspace{-.2in}
	\caption{\textbf{Evolution.} Evolution of training compressed 32 layer ResNet on Cifar100, using TRNs with different values of $r$ and the Tucker factorization method.}
	\label{convergence}
\end{figure}

\subsection{ResNet and Wide ResNet Compression}
Finally, we evaluate the performance of tensor ring nets  (TRN) on the Cifar10 and Cifar100 image classification tasks  \cite{krizhevsky2009learning}. Here, the input images are colored, of size $32 \times 32 \times 3$, belonging to 10 and 100 object classes respectively. Overall there are  50000 images for training and 10000 images for testing. 

{Table \ref{ana_resnet_table} gives the dimensions of ResNet before and after compression. A similar reshaping scheme is used for WideResNet. Note that for ResNet, we have compression gain for any $r \leq 22$; for WideResNet this bound is closer to $r\leq 150$, suggesting high compression potential.}


The results are given in Table \ref{resnet_cifar} demonstrates that TRNs are able to significantly compress both ResNet and WideResNet for both tasks.
Picking $r=10$ for TRN on ResNet  gives the same compression ratio as the Tucker compression method~\cite{kim2015compression}, but with almost 3\% performance lift on Cifar10 and almost 10\% lift on Cifar 100.
Compared to the uncompressed model, we see only a 2\% performance degradation on both datasets.

The compression of WideResNet is even more successful, suggesting that TRNs are well-suited for these extremely overparametrized models. At a $243\times$ compression TRNs give a better performance on Cifar10 than uncompressed ResNet (but with fewer parameters) and only a 2\% decay from the uncompressed  WideResNet. For Cifar100, this decay increases to 8\%, but again TRN of WideResNet achieves lower error than uncompressed ResNet, with overall fewer parameters.
Compared against the Tucker compression method \cite{kim2015compression}, at $5\times$ compression rate TRNs incur only 2-3\% performance degradation on both datasets, while Tucker incurs 5\% and 11\% performance degradation. The compressibility is even more significant for WideResNet, where to achieve the same performance as Tucker \cite{kim2015compression} at $5\times$ compression, TRNs can compress up to $243 \times$ on Cifar10 and $286\times$ on Cifar100. The tradeoff is runtime; we observe the Tucker model trains at about 2 or 3 times faster than TRNs for the WideResNet compression. However, for memory-constrained devices, this tradeoff may still be desirable.
\begin{table*}[t!]
	\begin{center}
		\begin{tabular}{ |c|c|c|c|c| }
			\hline
			&\multicolumn{2}{c|}{Uncompressed dims.} & \multicolumn{2}{c|}{TRN dimensions}\\
			\hline
			layer 		&shape 							& \# params 		 	&shape of composite tensor 							&  \# params 		\\
			\hline
			conv1		&$3\times 3 \times 3 \times 16$ 		& 432 			&		 $9\times 3 \times (4 \times 2 \times 2)$	  				& $20r^2$			\\
			unit1			&ResBlock(3, 16, 16)				& 4608			& 		 $9\times (4 \times 2 \times 2) \times (4 \times 2 \times 2)$	& $50r^2$			\\
			&ResBlock(3, 16, 16) $\times$ 4		& 18432			& 		 $9\times (4 \times 2 \times 2) \times (4 \times 2 \times 2)$	& $200r^2$		\\
			unit2			&ResBlock(3, 16, 32) 				& 13824			& 		 $9\times (4 \times 2 \times 2) \times (4 \times 4 \times 2)$	& $56r^2$			\\
			&ResBlock(3, 32, 32) $\times$ 4		& 73728			& 		 $9\times (4 \times 4 \times 2) \times (4 \times 4 \times 2)$	& $232r^2$		\\
			unit3			&ResBlock(3, 32, 64) 				& 55296			& 		 $9\times (4 \times 4 \times 2) \times (4 \times 4 \times 4)$	& $64r^2$			\\
			&ResBlock(3, 64, 64) $\times$ 4		& 294912			& 		 $9\times (4 \times 4 \times 4) \times (4 \times 4 \times 4)$	& $264r^2$		\\
			fc1			&64 $\times$ 10					& 650			&		 $(4 \times 4 \times 4)  \times 10$						& $22r^2$			\\
			\hline
			Total			&-								& 0.46M					&-												& $908r^2$			\\
			\hline
		\end{tabular}
	\end{center}
	\vspace{-.2in}
	\caption{{\textbf{Large convolution compression.} Dimensions of 32 layer ResNes on Cifar10 dataset. Each ResBlock($p$,$I$,$O$)
		includes a sequence: input  $\to$  Batch Normalization $\to$ ReLU $\to$ $p\times p\times I\times O$ convolution layer $\to$ Batch Normalization $\to$ ReLU $\to$ $p\times p\times O\times O$ convolution layer.  
		The input of length $I$ is inserted once at the beginning and again at the end of each unit. See  \cite{he2016deep} for more details.}}
	\label{ana_resnet_table}
\end{table*}

\begin{table*}[t!]
\begin{tabular}{ |l|c|c|c|c|c|c| }
 \hline

  &\multicolumn{3}{c|}{Cifar10} 
  &\multicolumn{3}{c|}{Cifar100}  \\\hline
  					Method	& Params 		&CR  & Err \% 					& Params 		&CR  	& Err \% \\
  \hline
 ResNet(RN)-32L 	 			& 0.46M  		 &$1\times$ 		&  7.50\cite{tensorflow} 	& 0.47M  		 &$1\times$ 		& 31.9 \cite{tensorflow}\\	
 Tucker-RN \cite{kim2015compression}				
 							& 0.09M		& $5\times$		& 12.3				& 0.094M 		&$5\times$		&42.2\\
 TT-RN($r=13$) \cite{garipov2016ultimate,novikov2015tensorizing}		
  							& 0.096M		& $4.8\times$		& 11.7				& 0.102M 		&$4.6\times$		& 37.1\\
 TRN-RN ($r=2$)		  		& 0.004M		&$115\times$ 		& 22.2 				& 0.012M		& $39\times$ 		& 51.3\\
 TRN-RN ($r=6$)		 		& 0.03M 		&$15\times$ 		& 19.2  				& 0.041M		&$12\times$ 		& 36.6 \\
 TRN-RN ($r=10$)				& 0.09M 		& $5\times$ 		& 9.4   				& 0.097M 		&$5\times$		& 33.3  \\
 \hline
 WideResNet(WRL)-28L  	 	& 36.2M   		&$1\times$ 		&  5.0 \cite{tensorflow}	&  36.3M 		&$1\times$		&  21.7 \cite{tensorflow}\\ 
 Tucker-WRN \cite{kim2015compression}				& 6.7M		& $5\times$		& 7.8 				& 6.7M 		& $5\times$		& 30.8\\
   TT-RN($r=13$) \cite{garipov2016ultimate,novikov2015tensorizing}			
   						& 0.18M	        &$201\times$		&  8.4				& 0.235M		&$154\times$		&  31.9 \\
   TRN-WRN ($r=2$)	  		& 0.03M		&$1217\times$		& 16.3 				& 0.087M		&$417\times$		&  43.9 \\
   TRN-WRN ($r=6$) 		& 0.07M 		&$521\times$		&  9.7  				& 0.126M		&$286\times$		&  30.3 \\
 TRN-WRN ($r=10$)			& 0.15M 		&$243\times$		&  {\bf 7.3}  			& 0.21M		&$173 \times$		&  28.3\\
   TRN-WRN(r=15)	 		& 0.30M 		&$122\times$ 		& {\bf 7.0}				&0.36M		&$100\times$		&  25.6\\
 \hline
\end{tabular}
\centering
\vspace{-.1in}
 \caption{\textbf{Large convolution results.} 32-layer ResNet (first 5 rows) and 28-layer Wide-ResNet (last 4 rows) on Cifar10 dataset and Cifar100 dataset, trained to 200 epochs, using a minibatch size of 128. 
 {The model is trained using SGD with momentum 0.9 and a decaying learning rate.} 
 CR = Compression ratio.}
 \label{resnet_cifar}
 \vspace{-.1in}
 \end{table*}
 
\vspace{-.1in}
\paragraph{Evolution}
Figure \ref{convergence}  shows the train and test errors during training of compressed ResNet on the Cifar100 classification task, for various choices of $r$ and also compared against Tucker tensor factorization. 
In particular, we note that the generalization gap (between train and test error) is particularly high for the Tucker tensor factorization method, while for TRNs (especially smaller values of $r$) it is much smaller. And, for $r = 10$, both the generalization error and final train and test errors improve upon the Tucker method, suggesting that TRNs are easier to train.

\label{sec:exp}



\section{Conclusion}
We have introduced a tensor ring factorization approach to compress deep neural networks for resource-limited devices. 
This is inspired by previous work that has shown tensor rings to have high representative power in image completion tasks. 
Our results show significant compressibility using this technique, with little or no hit in performance on benchmark image classification tasks.

One area for future work is the reduction of computational complexity. Because of the repeated reshaping needs in both fully connected and convolutional layers, there is  computational overhead, especially when $r$ is moderately large. This tradeoff is reasonable, considering our considerable compressibility gains, and is appropriate in memory-limited applications, especially if training is offloaded to the cloud. Additionally, we believe that the actual wall-clock-time will decrease as tensor-specific hardware and low-level routines continue to develop--we observe, for example, that numpy's \texttt{dot} function is considerably more optimized than Tensorflow's \texttt{tensordot}. Overall, we believe this is a promising compression scheme and can open doors to using deep learning in a much  more ubiquitous computing environment.
\label{sec:conclude}

	\bibliographystyle{ieee}
	\bibliography{Ref}

\begin{appendices}
\section{Merge ordering}\label{app:merge}
\begin{figure*}
	\centering
	\begin{subfigure}[b]{\textwidth}
		\includegraphics[width=.75\textwidth]{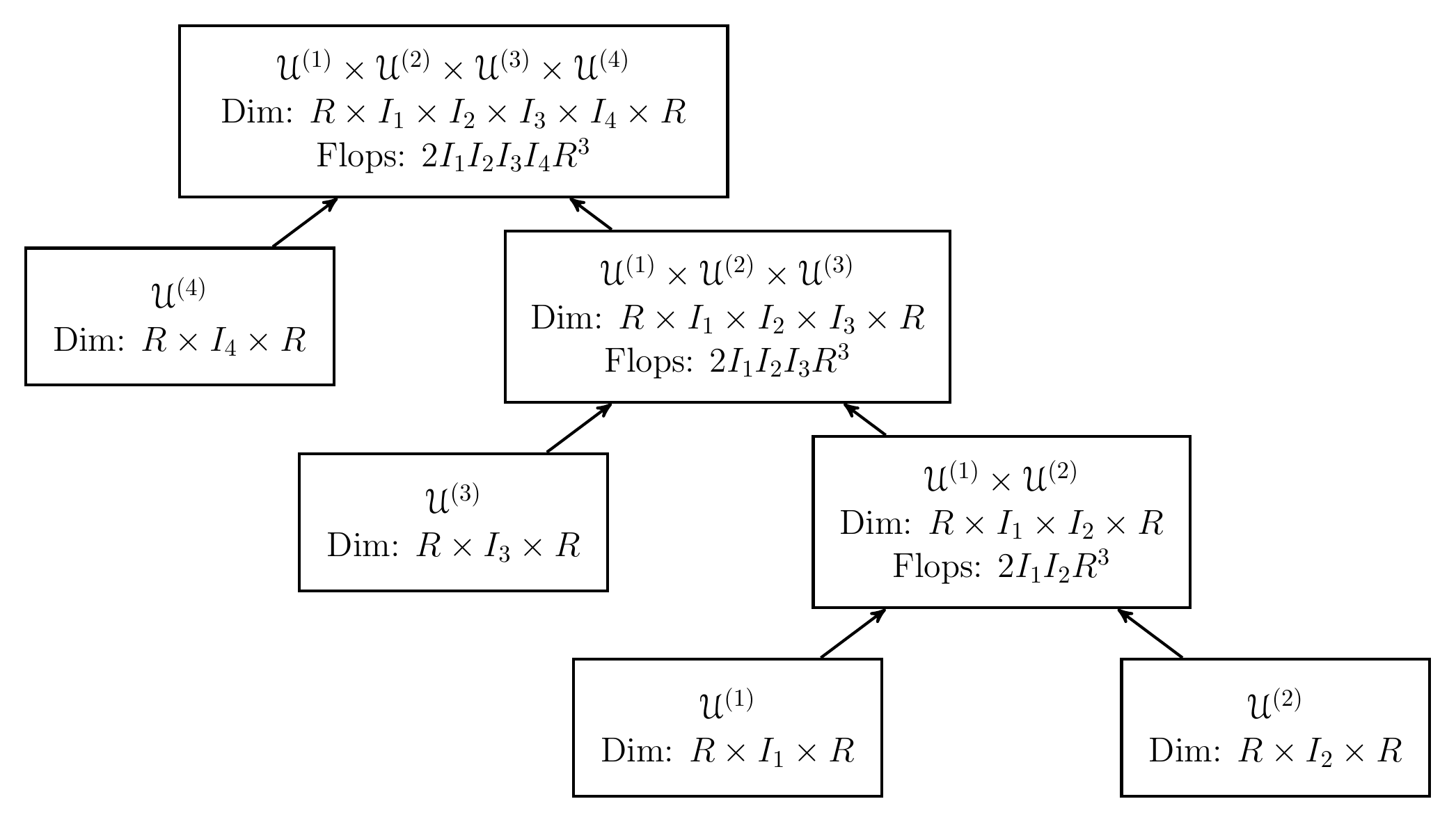}
		\centering
		\caption{Sequential merging}
		\label{ST}
	\end{subfigure}
	\begin{subfigure}[b]{\textwidth}
		\includegraphics[width=.75\textwidth]{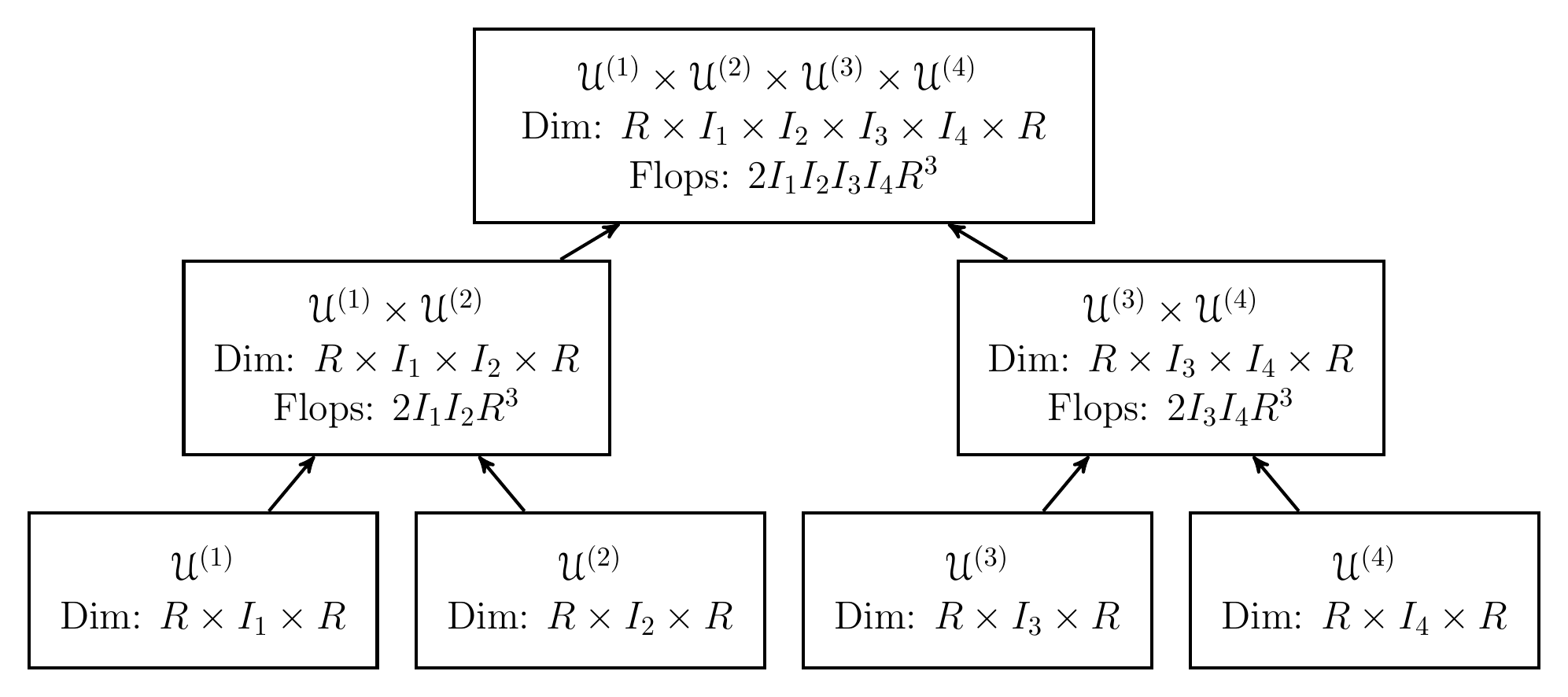}
		\centering
		\caption{Hierarchical merging}
		\label{HT}
	\end{subfigure}
	\caption{Merge ordering for a $4$th order tensor ring segment of shape $R\times I_1 \times I_2 \times I_4 \times I_4\times R$, with tensor ring rank $R$. In each node from top to bottom are tensor notation, tensor shape, and flops to obtain the tensor.}
\end{figure*}

For a \merge~operation, the order that each $\U^{(i)}$ is merged determines the total flop count and memory needs.
When $d$ is small, a \emph{sequential merging} is commonly applied.
However, when $d$ is large, we propose a \emph{hieratical merging} approach instead. 
For instance, Figures \ref{ST} and \ref{HT} show the two merge orderings when $d=4$, arriving at a total of  $2I_1I_2R^3 +2I_1I_2I_3R^3 + 2I_1I_2I_3I_4R^2$ flops to construct $\mathscr{U}^{(1,2,3,4)}$ using a sequential ordering, and
$2I_1I_2R^3+2I_3I_4R^3 +2 I_1I_2I_3I_4R^2$ flops using a hierarchical ordering.
To see how both methods scale with $I_k$ and $d$, if and $d = 2^D$, then a sequential merging gives
 and  Both quantities are upper bounded by $4R^3\tilde I^d$ which is a factor of $4R^3$ times the total degrees of freedom.

We can generalize this analysis by proving  theorem \ref{th-mergeordering}.
\begin{proof}
	\begin{enumerate}
		\item 
		Define $\tilde I = I_1=\cdots=I_d$.
		Any merging order can be represented by a binary tree. Figures \ref{ST} and \ref{HT} show the binary trees for sequential and hierarchical merging; note that they do not have to be balanced, but every non-leaf node has exactly 2 children. Each $U^{(i)}$ corresponds to a leaf of the tree. 
		
		To keep the analysis consistent, we can say that the computational cost of every leaf is 0 (since nothing is actually done unless tensors are merged).
		
		At each parent node, we note that the computational cost of merging the two child nodes is at least 2 $\times$ that required in the sum of both child nodes. 
		This is trivially true if both children of a node are leaf nodes. 
		For all other cases, define $D$ the number of leaf node descendents of a parent node. Then the computational cost at the parent is $2R^3 \cdot \tilde I^D$.
		If only one of the two child nodes is a leaf node, then
		we have a recursion
		\[
		2R^3 \cdot \tilde I^D = 2R^3 \tilde I\cdot  \tilde I^{D-1} \geq 4R^3 (\tilde I^{D-1})
		\]
		which is always true if $\tilde I \geq 2$.
		If both children are not leaf nodes, then define  $D_1$, and $D_2$  the number of leaves descendant of  two child nodes, with $D= D_1 + D_2$. Then the recursion is
		\[
		2R^3 \cdot \tilde I^D = 2R^3 \tilde I^{D_1} \tilde I^{D_2} \geq 4R^3 (\tilde I^{D_1}+\tilde I^{D_2})
		\]
		where the bound is always true for $\tilde I \geq 2$ and $D_1, D_2 \geq 2$.
		Note that every non-leaf node in the tree necessarily has two children, it can never be that $D_1=1$ or $D_2 = 1$. 
		
		The cost of merging at the root of the tree is always $2R^3 \tilde I^d = 2R^3 I$. Since each parent costs at least $2\times$ as many flops as the child, the total flop cost must always be between $2R^3I$ and $4R^3I$.
		
		\item For the storage bound, the analysis follows from the observation that the storage cost at each node is $R^2 \tilde I^D$, where $D$ is the number of leaf descendants. Therefore if $\tilde I \geq 2$, the most expensive storage step will always be at the root, with $R^2(\tilde I^{d_1} + \tilde I^{d_2} + \tilde I^{d})$ storage cost, where $d = d_1 + d_2$ for any partition. Clearly, this value is lower bounded by $R^2 \tilde I^d = R^2I$. And, for any partition $d_1 + d_2 = d$, for $\tilde I \geq 2$, it is always $\tilde I^{d_1} + \tilde I^{d_2} \leq \tilde I^d$. Therefore the upper bound on storage is $2 R^2 \tilde I^d = 2R^2 I$.
		
		\item It is sufficient to show that for any $d$ power of 2, a sequential merging is more costly in flops than a hierarchical merging, since anything in between has either pure sequential or pure hierarchical trees as subtrees.
		
		Then a sequential merging gives $2R^3\sum_{i=2}^d \tilde I^i$ flops.
		If additionally $d = 2^D$ for some integer $D > 0$, then a hierarchical  merging costs
		$2R^3 \sum_{i=2}^{D} 2^{D-i}\tilde I^{2^i}$ flops. To see this, note that in a perfectly balanced binary tree of depth $D$, at each level $i$ there are $2^{D-i}$ nodes, each of which are connected to $2^i$ leaves.
		
		We now use induction to show that whenever $d$ is a power of 2, hierarchical merging (a fully balanced binary tree) is optimal in terms of flop count.
		If $d=2$, there is no variation in merging order.
		Taking $d = 4$, a sequential merging costs $2R^3 (\tilde I^3 + \tilde I^3 + \tilde I^4)$ and a hierarchical merging costs
		$2R^3(2 \tilde I^2 + \tilde I^4)$, which is clearly cheaper. 
		For some $d$ a power of 2, define $S$ the cost of sequential merging and $H$ the cost of hierarchical merging. Define $G=2R^3\tilde I^{2d}$ the cost at the root for any binary tree with $2d$ leaf nodes. (Note that the cost at the root is agnostic to the merge ordering.)
		Then for $\hat d = 2d$, a hierarchical merging costs
		$2H + G$ flops.
		The cost of a sequential merging is 
		\begin{eqnarray*}
		S + 2R^3\tilde I^d\sum_{i=1}^{ d}\tilde I^i &=& S+2R^3\tilde I^{d-1}\sum_{i=2}^d \tilde I^i + G\\
		&=& S + S\tilde I^{d-1} + G - 2R^3 d.
		\end{eqnarray*}
	Since $2R^3 d$ is the cost at the root for $d$ leafs, $S > 2R^3 d$, and therefore the above quantity is lower bounded by $G + \tilde I^{d-1}S$, which for $d \geq 2$ and $\tilde I \geq 2$, is lower bounded by $G + 2S$. By inductive hypothesis, $S > H$, so the cost of sequential merging is always more than that of hierarchical merging, whenever $d$ is a power of 2.
	
	\end{enumerate}

%
 \end{proof}


\section{Initialization}\label{ini_proof}
If $x$ and $y$ are two independent variables, then $\text{Var}[xy] = \text{Var}[x]\text{Var}[y] + \text{Var}[x](\mathbb{E}[y])^2 + \text{Var}[y](\mathbb{E}[x])^2$ \cite{prob}.
Thus a product of two independent symmetric distributed random variables with mean $0$ and variance $\sigma^2$ itself is symmetric distributed with mean $0$ and variance $\sigma^4$ (not Gaussian distribution).
Further extrapolating, in a matrix or tensor product, each entry is the summation of $R$ independent variables with the same distribution. The central limit theorem gives that the sum can be approximated by a Gaussian $\mathcal{N}(0, R\sigma^4)$ for large $R$.
Thus if all tensor factors are drawn i.i.d. from $\mathcal{N}(0, \sigma^2)$, then after merging $d$ factors the merged tensor elements will have mean $0$ and variance $R^d \sigma^{2d}$.

\end{appendices}

\end{document}